\documentclass[draft]{agujournal2019}
\usepackage{url} 
\usepackage{lineno}
\usepackage[inline]{trackchanges} 
\usepackage{soul}
\usepackage{amsmath} 

\draftfalse

\usepackage{float}
\begin{document}

\title{How to systematically develop an effective AI-based bias correction model?}

\authors{Xiao Zhou\affil{1}\textsuperscript{†}, Yuze Sun\affil{1}\textsuperscript{†}, Jie Wu\affil{2}, Xiaomeng Huang\affil{1}}

\affiliation{1}{Department of Earth System Science, Ministry of Education Key Laboratory for Earth System Modelling, Institute for Global Change Studies, Tsinghua University, Beijing, China}
\affiliation{2}{State Key Laboratory of Climate System Prediction and Risk Management/China Meteorological Administration Key Laboratory for Climate Prediction Studies, National Climate Centre, China Meteorological Administration, Beijing, 100081}

\correspondingauthor{Xiaomeng Huang}{hxm@tsinghua.edu.cn}
\footnotetext{\textdagger These authors contributed equally to this work.}

\begin{abstract}
This study introduces ReSA-ConvLSTM, an artificial intelligence (AI) framework for systematic bias correction in numerical weather prediction (NWP). We propose three innovations by integrating dynamic climatological normalization, ConvLSTM with temporal causality constraints, and residual self-attention mechanisms. The model establishes a physics-aware nonlinear mapping between ECMWF forecasts and ERA5 reanalysis data. Using 41 years (1981–2021) of global atmospheric data, the framework reduces systematic biases in 2-m air temperature (T2m), 10-m winds (U10/V10), and sea-level pressure (SLP), achieving up to 20\% RMSE reduction over 1–7 day forecasts compared to operational ECMWF outputs. The lightweight architecture (10.6M parameters) enables efficient generalization to multiple variables and downstream applications, reducing retraining time by 85\% for cross-variable correction while improving ocean model skill through bias-corrected boundary conditions. The ablation experiments demonstrate that our innovations significantly improve the model's correction performance, suggesting that incorporating variable characteristics into the model helps enhance forecasting skills.
\end{abstract}

\section{Introduction}

Numerical weather prediction (NWP) is crucial in weather forecasting, providing indispensable guidance across temporal scales from nowcasting to seasonal forecasting \cite{bauer2015quiet}. As society becomes more dependent on accurate forecasts, there is an increasing demand for high-quality predictions, particularly in extreme events such as heat waves and cold surges, which can have severe social and economic impacts\cite{bras2023much,miao2024unveiling}. Furthermore, atmospheric forecasts serve as critical boundary conditions for coupled Earth system models, where their accuracy directly governs the predictive capabilities of oceanographic and cryospheric simulations through dynamic coupling mechanisms. While the ECMWF's Integrated Forecasting System (IFS) represents the state-of-the-art in global operational prediction \cite{molteni1996ecmwf}, persistent systematic biases still exist, which arise from three fundamental sources: (1) inadequate spatial resolution to resolve subgrid-scale processes \cite{mishra2021impact}, (2) inherent limitations in physical parameterization schemes \cite{berner2017stochastic,brenowitz2018prognostic}, and (3) uncertainties in initial/boundary condition specification \cite{peng2006effect}. Current bias correction paradigms predominantly employ statistical postprocessing techniques, including uni-variate regression frameworks \cite{turco2017bias}, adaptive filtering techniques \cite{chandramouli2022online}, and probabilistic calibration methods \cite{yumnam2022quantile}. However, these approaches exhibit two critical limitations: first, their grid-point-wise (local) implementations neglect spatially correlated (remote) error structures across multiple scales; second, their quasi-linear mathematical foundations prove inadequate for disentangling nonlinear interactions between dynamical and physical processes. Exploring novel approaches to improve bias correction performance is necessary to bridge these methodological gaps.

The emergence of deep learning has significantly improved the predictive capabilities in the atmosphere field. Foundation models for AI-driven weather and climate prediction \cite{pathak2022fourcastnet,bi2023accurate,xiong2023ai} now demonstrate effectiveness in capturing complex earth system interactions. This advancement has extended to bias correction through data-driven models that learn bias patterns between model outputs and observational truths \cite{bengio2009learning,glorot2010understanding,tishby2015deep}. At synoptic timescales (\~ 7 days), recent innovations mainly focus on surface atmospheric variables, including 2-meter air temperature (T2m), 10-meter wind components (U10 and V10), mean sea level pressure (SLP), and precipitation rate (Pr). \citeA{han2021deep} developed a UNet\cite{ronneberger2015u} architecture for 2-meter air temperature anomaly correction, demonstrating better performance than traditional anomaly-based numerical optimization methods. \citeA{zhang2023deep} and \citeA{fang2023short} implemented a Gated Recurrent Unit (GRU)-based system \cite{dey2017gate} for real-time sea surface wind adjustment, performing well under normal conditions and extreme events such as typhoons. \citeA{wang2022customized} and \citeA{mcgibbon2024global} applied the GAN \cite{goodfellow2020generative} model for bias correction of precipitation, effectively reducing the short-term and cumulative bias in precipitation rate. While these AI-driven approaches make significant progress, some fundamental limitations persist: (1) Most studies are limited to local bias correction of variables and fail to provide a global perspective for overall correction; (2) Models are designed and optimized for specific variables (e.g., T2m), demonstrating poor generalization to multivariate correction tasks; (3) Current model architectures insufficiently incorporate atmospheric characteristics, potentially violating some laws during error adjustment.

To fundamentally enhance the AI-based bias correction effect,  we argue that the intrinsic physical characteristics of the corrected variables should be fully considered during the data pre-processing and model design phases. This approach would better leverage the AI model's ability to learn nonlinear relationships, further discovering the evolution patterns of biases over time and space, thereby enhancing the correction effectiveness. The model should also be lightweight, efficient, and capable of generalization. The rest of this article is structured as follows. Section 2 presents the data and methods used in this study. Section 3 will take the 2-meter air temperature (T2m) as an example to analyze its intrinsic physical characteristics and provide a complete introduction to the establishment process of the AI correction model. Section 4 shows the model performance, the ablation experiment, and downstream tasks. Finally, Section 5 provides the conclusion and discussion.

\section{Data and Methods}

\subsection{Data}

This study utilizes seasonal forecast data from the European Centre for Medium-Range Weather Forecasts (ECMWF), providing seasonal outlooks of Earth system evolution. The original 1°×1° (180×360 grid) ensemble forecasts were initialized every 6 hours during 1981-2021, comprising 25-member ensembles \cite{johnson2019seas5}. Data were aggregated into daily averages, with the 25-member ensemble mean computed for each temporal instance. ERA5 reanalysis data \cite{hersbach2020era5} served as the ground-truth benchmark, maintaining spatiotemporal consistency with the forecast data. While 2-m air temperature (T2m) constituted the primary training variable, 10-m zonal wind (U10), 10-m meridional wind (V10), and mean sea-level pressure (SLP) were employed to assess cross-variable generalization capabilities. To evaluate downstream applicability, the AI-driven Global Ocean Modeling System (AI-GOMS) \cite{xiong2023ai} was implemented, with forecast outputs serving as boundary conditions and Hybrid Coordinate Ocean Model (HYCOM) \cite{chassignet2007hycom} reanalysis data as validation targets. Key oceanic variables, including sea surface temperature (SST) and zonal current velocity (U), were selected.

\subsection{Methods}

This study employs a neural network model for forecast bias correction. The architecture integrates multiple foundational components: convolutional layers extract spatial error characteristics \cite{lecun1998gradient}, and long short-term memory (LSTM) modules capture temporal error evolution patterns \cite{shi2015convolutional}. At the same time, residual connections \cite{he2016deep} and self-attention blocks \cite{lin2020self} are used to enhance correction performance. Strategically implemented batch normalization layers \cite{ioffe2015batch} between network stages stabilize gradient flow and optimize model convergence. The dataset was partitioned using a decadal sampling strategy, with five non-consecutive years (1981, 1991, 2001, 2011, 2021) selected at ten-year intervals as a testing set, while the remaining years constituted the training set. This stratified sampling approach rigorously evaluates model correction performance across distinct climate phases rather than single-year calibration scenarios. To assess model correction efficacy, we adopted the method from prior studies \cite{pathak2022fourcastnet,cao2025ai}, employing Root Mean Squared Error (RMSE) (Equation \ref{eq:rmse}) and Anomaly Correlation Coefficient (ACC) (Equation \ref{eq:acc}) as quantitative evaluation metrics.

The computational implementation utilizes an NVIDIA A100 GPU for model training and evaluation. The backbone model achieves convergence within 100 epochs (approximately 2 hours) with a maximum batch size of 4 constrained by GPU memory. The model contains 10,648,834 trainable parameters optimized using the Adam algorithm with mean squared error (MSE) loss. Based on the backbone model, we further fine-tune parameters using other variables to evaluate its cross-variable generalization capability.

\section{Physical consideration}

\begin{figure}
\noindent\includegraphics[width=\textwidth]{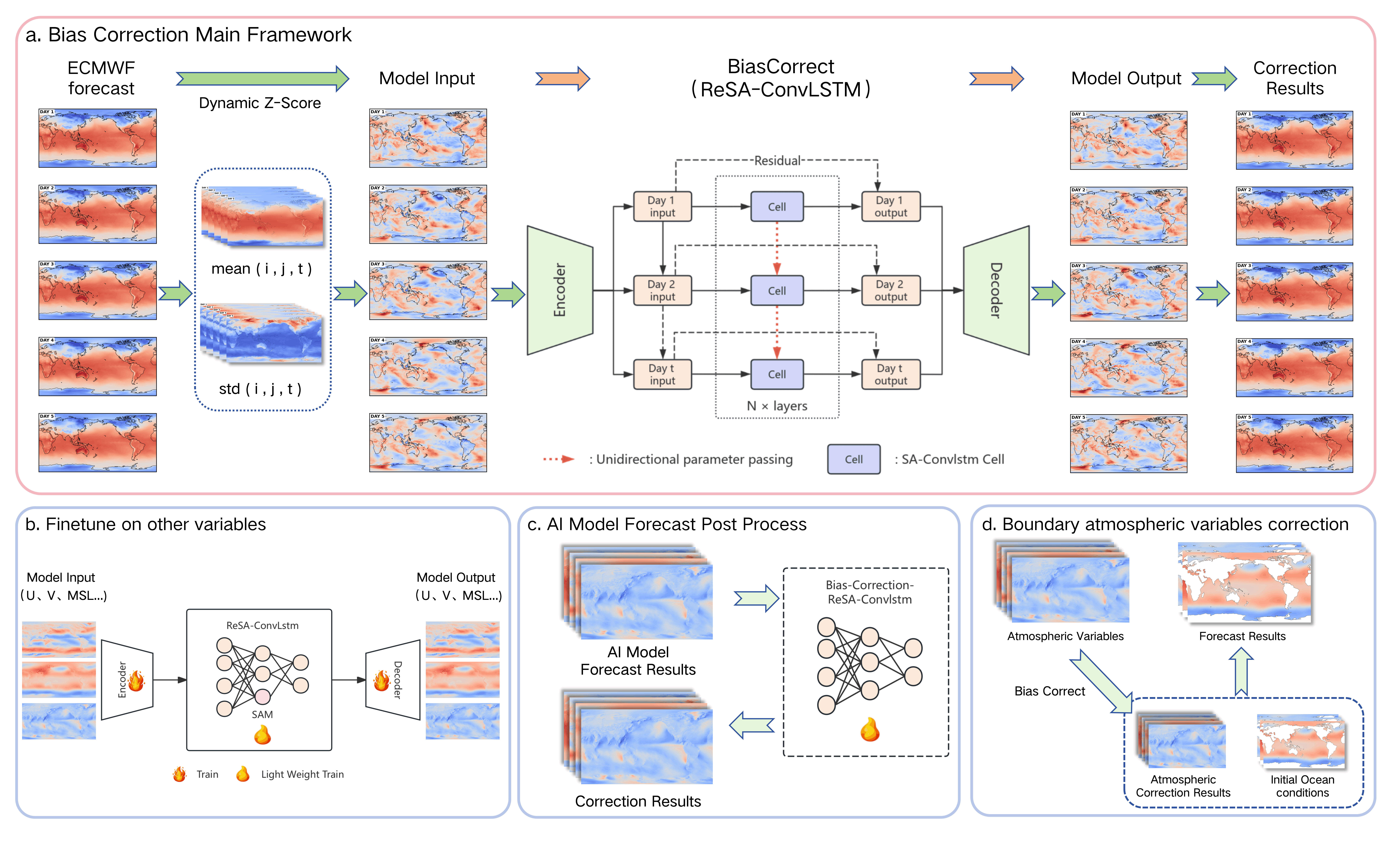}
\caption{An overview of the proposed bias correction architecture. (a) Overview of the core architecture illustrating the bias correction mechanism.  Three distinct downstream applications are integrated with the framework: (b) Parameter-adaptive fine-tuning for cross-variable optimization by leveraging the pre-trained backbone model, (c) Output-level correction through direct rectification of AI-model forecast results, and (d) Input-level correction through atmospheric variables serving as oceanic boundary conditions.}
\label{framework_diagram}
\end{figure}

This section explicitly outlines our model design philosophy. A robust AI correction framework necessitates three major components: (1) a data pre-processing process, (2) core architecture selection, and (3) improvement and optimization for model structure. We posit that effective bias correction models must consider the intrinsic variables during development. This approach can significantly facilitate the capture capability of nonlinear bias features, enhance learning convergence rates, and optimize correction performance. Here, we use T2m as an example to introduce the model development process.

Data normalization, a critical pre-processing step in AI workflows, fundamentally governs model learning capacity through distributional reshaping. \citeA{glorot2010understanding,ioffe2015batch} demonstrates that AI training data is most beneficial when scaled into unit-variance ranges and follows a quasi-normal distribution. The climatology means and standard deviation of T2m reveal the significant contrast between the tropical and polar regions (Figure \ref{mean_std}a). Equatorial regions exhibit higher mean temperatures with smaller inter-annual variability, while polar regions are the opposite. The statistical analysis shows that the temperature is a significant non-Gaussian distribution. Conventional z-score normalization (Eq. \ref{eq:z-score}) employs static statistical parameters (\(\mu\))and (\(\sigma\)) to scale variables into unit-variance ranges, yet fails to neither mitigate persistent equator-to-pole temperature contrast nor alter the data distribution (Figure \ref{data_orgin_zscore_our}b). Our dynamic normalization framework implements grid-wise climatological means (\(\mu_{i,j,t}\)) and inter-annual variance (\(\sigma_{i,j,t}\)) to remove equatorial-polar differences. Meanwhile, the data distribution approximates normal distribution, enhancing the model's ability to learn bias characteristics (Figure \ref{data_orgin_zscore_our}c).

In selecting foundational model architectures, conventional correction systems predominantly employ UNet \cite{ronneberger2015u} frameworks. This architecture leverages symmetrical encoder-decoder structures with successive downsampling and upsampling operations to hierarchically extract spatiotemporal features. Convolutional kernels operate along the channel dimension within the encoder pathway to capture error propagation patterns across the temporal axis. For instance, when correcting day-3 forecasts, the model inherently incorporates temporal dependencies from both preceding (e.g. day-2) and subsequent (e.g. day-4) forecast results. However, the critical requirement for capturing temporal dependencies is preventing future information leakage into past time steps during feature extraction (Figure \ref{time_arrow}), as this could lead to causal relationship errors in time. Therefore, we adopt the Convolutional Long Short-Term Memory (ConvLSTM) \cite{shi2015convolutional} architecture as the foundational framework. This model effectively extracts spatial features while strictly preserving temporal causality, thereby preventing future information leakage into the past.

Building upon the foundational architecture, we further implement some refinements to optimize model performance. As demonstrated in Figure \ref{data_orgin_zscore_our}c (showing 1-5 Mar 2021 T2m forecasts), our dynamic normalization effectively mitigates spatial contrast but reveals some amplitude disparities: synoptic-scale temperature variations (~ \~0.1 normalized units) remain an order of magnitude smaller than background signals (~ \~1 normalized unit), creating gradient vanishing risks for neural networks. To resolve this, we integrate (1) temporally constrained self-attention mechanisms\cite{lin2020self} that adaptively amplify meteorologically significant patterns through position-aware weighting and (2) residual connections\cite{he2016deep} facilitate the training of deeper networks for nonlinear error feature extraction while mitigating gradient vanishing problems. These improvements can effectively extract small signal changes in the bias evolution process.

Based on the analysis above, our model is designed as Figure \ref{framework_diagram}. The model consists of the backbone and downstream tasks. In the backbone model (Figure \ref{framework_diagram}a), we employ a dynamic normalization method for data pre-processing and select ConvLSTM as the foundational architecture of the model. Building upon this, we introduce the self-attention block and residual block to extract small bias features further. The model is named as ReSA-ConvLSTM. Additionally, our model demonstrates excellent generalization capabilities and scalability. Slight adjustments to the model can be quickly applied to the correction of other variables (e.g., U10, V10, SLP) (Figure \ref{framework_diagram}b). The model can also serve as a plugin for other forecasting models, improving forecast skills (Figure \ref{framework_diagram}c). Furthermore, the corrected atmospheric forecast variables can be used as boundary conditions for ocean forecasting models (Figure \ref{framework_diagram}d). In Chapter 4, we will present the correction results and conduct a series of ablation experiments to evaluate the contributions of the innovations above. We will also illustrate their effects on generalization performance and related downstream tasks.

\section{Results}
\subsection{Model Performance}

\begin{figure}
\noindent\includegraphics[width=\textwidth]{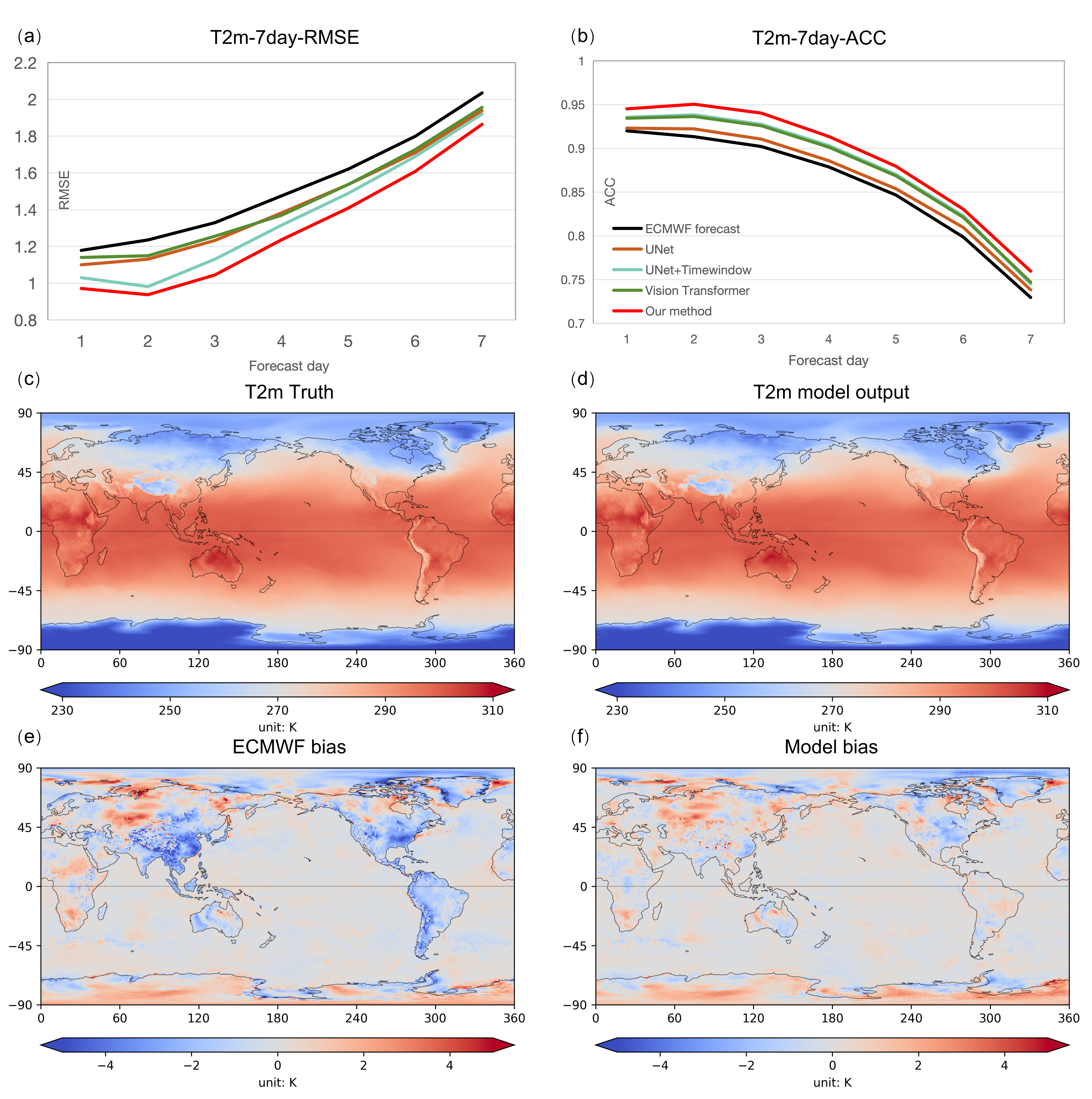}
\caption{
Evaluation of Bias Correction Performance.    (a) Temporal evolution of 2m temperature (T2m) Root Mean Square Error (RMSE; units: K) and (b) Anomaly Correlation Coefficient (ACC) across 7-day forecasts, comparing ECMWF predictions with four correction AI methods: UNet (orange), UNet-TimeWindow (blue), Vision Transformer (green), and our proposed method (red).    (c-f) Spatial validation analysis showing (c) ERA5 reanalysis ground truth, (d) bias-corrected predictions from our model, with corresponding error distributions for (e) raw ECMWF forecasts versus ERA5 and (f) corrected predictions versus ERA5.  All spatial maps represent composite means of 7-day forecasts initialized on the first day of each month during validation years (1981, 1991, 2001, 2011, 2021).}
\label{main_result}
\end{figure}

We train a 7-day correction model using the T2m variable (Figure \ref{T2m_3day_5day}) and quantitatively measure the bias correction performance. All forecast cycles were initialized throughout the experimental period on the first day of each calendar month. The average RMSE and ACC for each forecast are calculated over all months (Figure \ref{main_result} a-b). We also replicated the current three major AI correction models (UNet, UNet-TimeWindow, and Vision Transformer) and compared their correction performance. The corrected 7-day forecasts demonstrate better performance over ECMWF operational predictions and other AI models. Quantitative evaluation reveals ECMWF's RMSE ranges from 1.2 (day 1) to 2.1 (day 7), while our model maintains RMSE \textless1.0 (days 1-3) and \textless1.8 (days 4-7), achieving up to 20\% error reduction against ECMWF with consistent out-performance over comparative AI models.

Figure \ref{main_result}c shows the climatology mean T2m from the ground truth (ERA5), while Figure \ref{main_result}d presents the bias-corrected model outputs. The corrected fields can generally capture the mean state of observed T2m. Spatial bias composites (Figure \ref{main_result}e-f) further demonstrate effective mitigation of ECMWF's biases:  cold biases in East Asia/North America/South America (average underestimate 2.3°C corrected to 0.7°C) versus warm biases in Africa (average overestimate 2.5°C corrected to 1.7°C).  A case study through March 1-5, 2021  (Figure \ref{true_ec_model_bias}) demonstrates significant bias reduction, with comparative error distributions against UNet/UNet-TimeWindow/Vision Transformer baselines detailed in Appendix Figure \ref{ec_unet_vit_our_bias}. The significant decrease in RMSE, consistent improvement in ACC across all forecast days, and enhanced spatial skill collectively demonstrate our model's effectiveness in mitigating systematic biases inherent in the original forecasting system.

\subsection{Ablation Study}

\begin{figure}
\noindent\includegraphics[width=\textwidth]{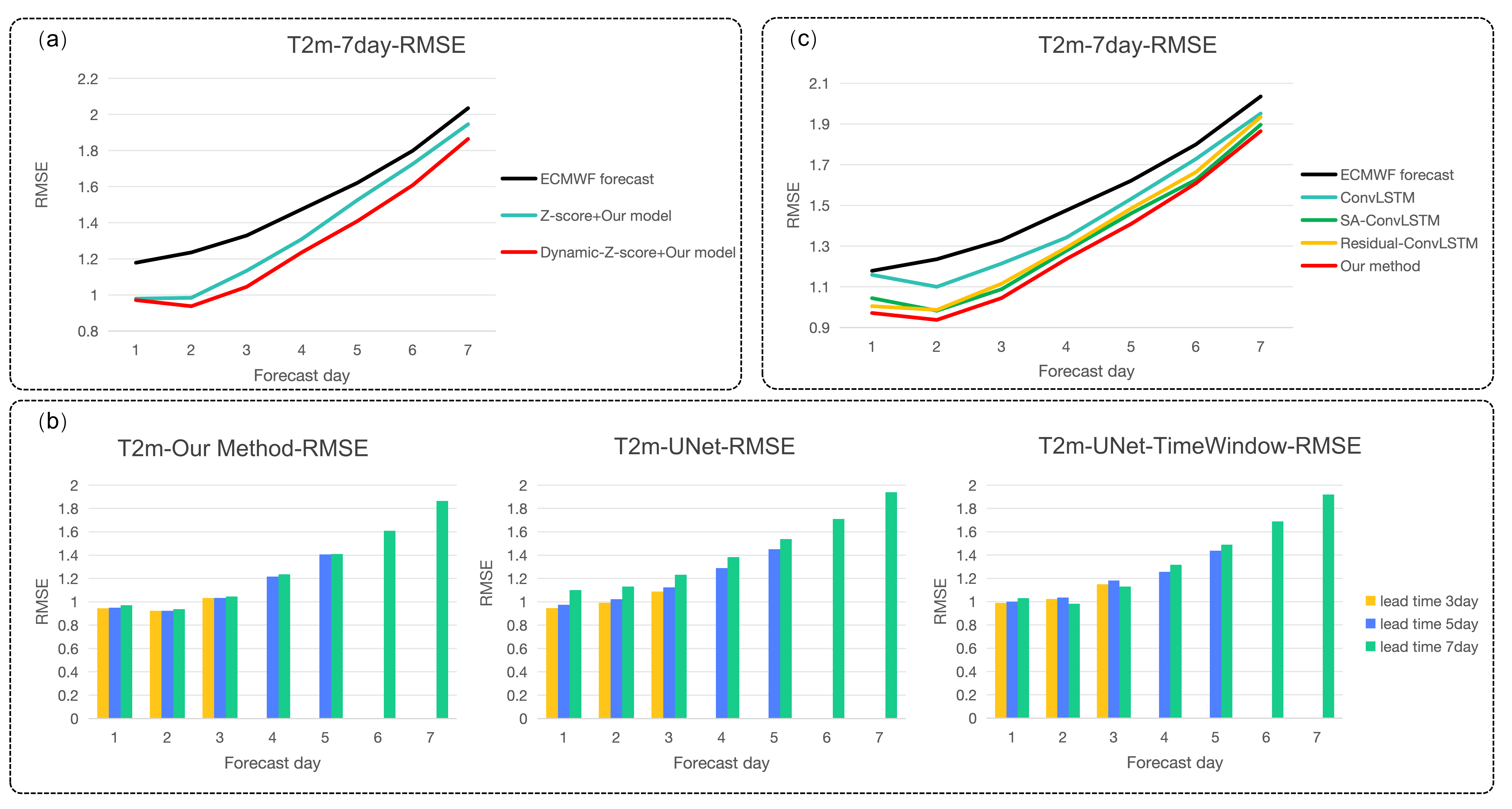}
\caption{Ablation experiment results. (a) Validation of the data processing method. The black curve shows the ECMWF forecast, the blue curve represents the Z-score normalization, and the red curve shows the spatiotemporal dynamic normalization (our method). The x-axis is forecast days, and the y-axis is the RMSE of T2m, with a 7-day forecast period. (b) Validation of the unidirectional time vector arrow. The three graphs compare the results of our method, UNet, and UNet-TimeWindow on the same dataset. The x-axis represents forecast lead time, and the y-axis shows T2m RMSE. Yellow, blue, and green bars indicate 3-/5-/7- day lead times. (c) Validation of the model architecture optimization. The curves show correction results for ECMWF (black), ConvLSTM (blue), SA-ConvLSTM (green), Residual-ConvLSTM (yellow), and our proposed ReSA-ConvLSTM (red).}
\label{Ablation_Study}
\end{figure}

Ablation studies systematically quantify the contributions of three architectural innovations to correction efficacy. Controlled experiments with identical model architectures demonstrate that our dynamic normalization method enhances forecast skill by 6.4\% beyond day 3 compared to conventional Z-score normalization (Figure \ref{Ablation_Study}a). These results explicitly confirm the critical sensitivity of correction performance to input normalization strategies, demonstrating the crucial role of data pre-processing in the bias correction AI model.

To assess architectural impacts on correction performance, we conducted multi-lead-time experiments (3-/5-/7-day), comparing our method with previous architectures (UNet, UNet-TimeWindow). Theoretically, employing varying forecast lead times for model training should not induce statistically significant differences in bias correction performance since the future result should not affect the past. For example, as shown in Figure \ref{Ablation_Study}b, the day-5 correction performance remains unchanged for our model with 5-day or 7-day lead-time training. In contrast, other architectures exhibit progressive error increases, implying the temporal contamination of future results to the past. The findings necessitate the principle that future information should not leak into the past. The model should not violate the causal relationship in time.

To assess architectural enhancements, we start from the basic architecture model (ConvLSTM) and add a self-attention block (SA-ConvLSTM), residual connections (Residual-ConvLSTM), and the combination of the above two (our model)  (Figure \ref{Ablation_Study}c). Using identical training schemes and datasets, experimental results demonstrate significant RMSE reductions compared to the basic architecture model, illustrating an improved correction performance. This is because introducing two blocks enhances the model's ability to extract small signal changes. Therefore, we use ReSA-ConvLSTM as our final model to correct the bias for the atmospheric variables.  This analysis indicates that it is necessary to optimize and modify the basic architecture based on the characteristics of atmospheric variables to meet the specific task requirements better.

\subsection{Downstream Tasks}

\begin{figure}
\noindent\includegraphics[width=\textwidth]{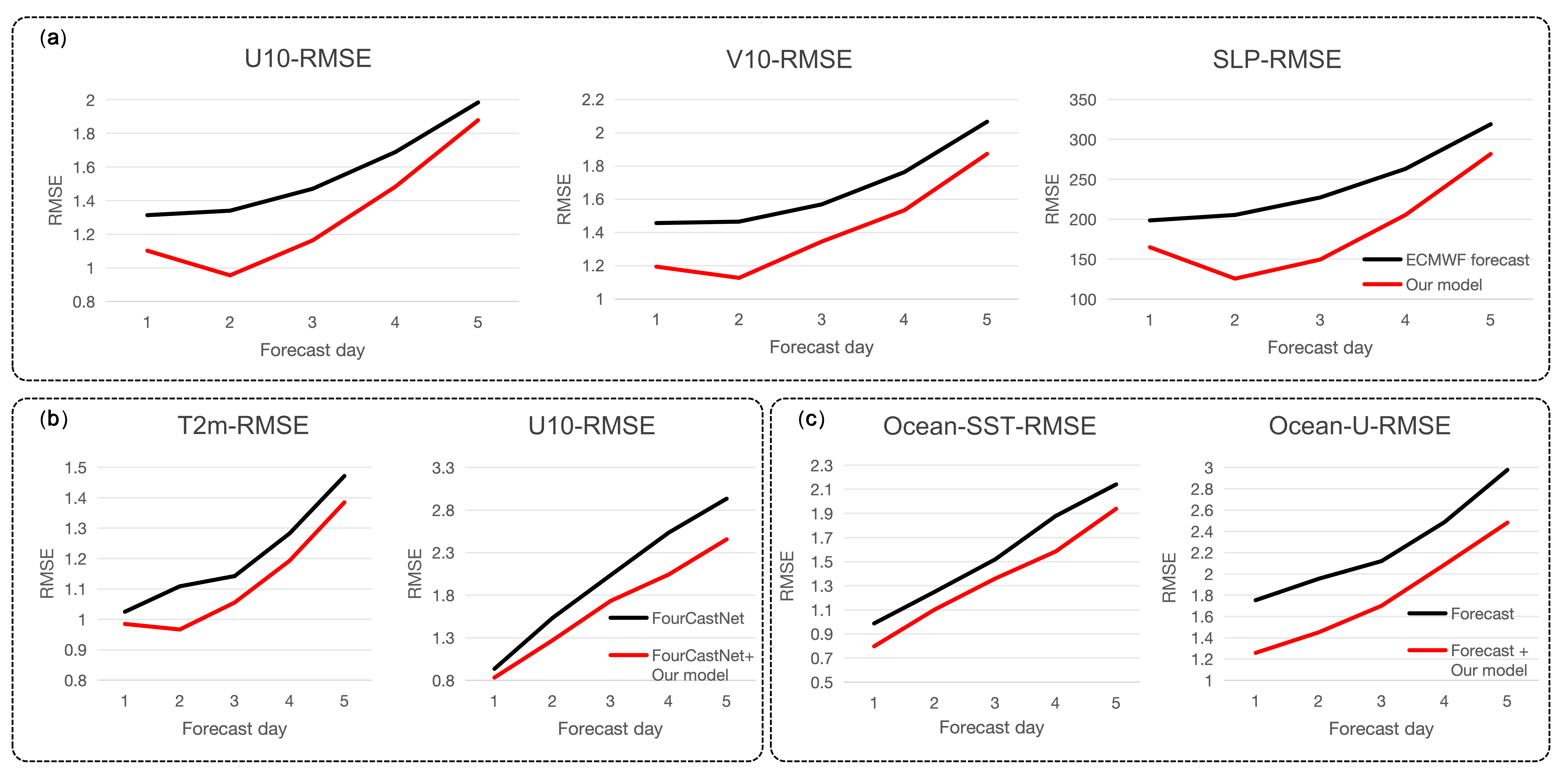}
\caption{Downstream tasks correction performance. (a) Correction performance of partially fine-tuned models on U10, V10, and SLP variables.  We adapt the T2m backbone model through parameter freezing and partial fine-tuning for downstream meteorological variables.  (b) Comparison between original AI-based weather forecasting outputs (black) and corrected predictions (red) using our method.  (c) Ocean variable predictions (SST and U) generated by feeding corrected atmospheric forcing fields (T2m, U10, and V10) into the ocean model.  Black curves in all panels represent baseline predictions from ECMWF or original AI models, while red curves show corrected results after our pre/post-processing.}
\label{downstream_task}
\end{figure}

An effective AI correction framework should demonstrate cross-variable adaptability and operational flexibility. For cross-variable generalization, models initialized with T2m-trained weights achieve convergence within 20 minutes (85\% faster than full training) for U10, V10, and SLP variables and show a significant correction effect (Figure \ref{downstream_task}a).  It demonstrates our model's capacity to extend bias correction to diverse atmospheric variables with minimal training expense.  Our model enables seamless integration as pre-/post-processing modules for numerical and AI-based forecasts. When deployed as a modular plugin within AI forecasting models (we use FourCastNet \cite{pathak2022fourcastnet} as an example), the system enables further bias correction, enhancing forecast skill by about 10\% (Figure \ref{downstream_task}b). When coupled with ocean models through corrected atmospheric boundary conditions, significant forecast improvement emerges (Figure \ref{downstream_task}c). These experiments confirm the good performance in different downstream tasks.

\section{Conclusions and Discussion}

This study presents a novel AI-driven framework (ReSA-ConvLSTM) for systematic bias correction in weather forecast. By integrating dynamic climatological normalization, ConvLSTM with temporal causality constraints, and residual self-attention mechanisms, the model achieves significant improvements over conventional statistical postprocessing and existing AI methods. Key results demonstrate an up to 20\% reduction in RMSE across 1–7 day forecasts for 2-m air temperature (T2m), with enhanced generalization to wind components (U10/V10) and sea-level pressure (SLP). The lightweight architecture (10.6M parameters) enables efficient downstream task adaptation, reducing retraining time by 85\% for cross-variable correction and improving ocean model forecast skill through bias-corrected boundary conditions. These advancements highlight the potential of physics-aware AI systems to enhance forecast skills.

Current AI-driven correction models predominantly rely on data-driven paradigms, probably overlooking fundamental meteorological principles. For instance, UNet-based frameworks frequently violate temporal causality by indiscriminately blending past-future features. This work provides a critical insight that tailoring AI architectures to variable-specific physical characteristics is essential for effective bias correction. This methodology not only outperforms pure data-driven counterparts but also reduces model complexity. Furthermore, integrating physical processes into the AI model design may help open the model's black box, improve its physical interpretability, and aid researchers in identifying potential forecasting factors that influence predictability. Our study further confirms that meteorologically conscious AI design can unlock more significant potential than generic architectures in the atmosphere field.

While the proposed framework demonstrates robust bias correction capabilities at synoptic timescales (1–7 days), extending its effectiveness to subseasonal-to-seasonal (S2S) predictions remains a critical challenge. Our experiments reveal a marked decline in correction efficacy beyond 7-day lead times relative to short-term forecasts. This limitation arises from two interconnected factors: First, the inherent predictability barrier of chaotic atmospheric systems, where initial-condition uncertainties increasingly dominate at extended lead times, and second, current AI architectures' limited capacity to capture slow-varying climate modes (e.g., MJO, ENSO) that fundamentally govern S2S predictability. Addressing how to synergistically integrate physical mechanisms with AI models through enhanced data preprocessing, tailored architectures, and optimized training strategies—while simultaneously extending forecast horizons and improving predictive skill—remains an urgent challenge requiring prioritized investigation.

%
%

\section*{Open Research}

{Data Availability Statement}
The ECMWF seasonal forecast data used in this study are available through the European Centre for Medium-Range Weather Forecasts’ (ECMWF) Meteorological Archival and Retrieval System (MARS) under institutional license agreements (https://www.ecmwf.int) \cite{johnson2019seas5}. ERA5 reanalysis data are publicly accessible via the Copernicus Climate Data Store (https://cds.climate.copernicus.eu) \cite{hersbach2020era5}.

\bibliography{ref}

\begin{thebibliography}{}

\bibitem [\protect \citeauthoryear {%
Bauer%
, Thorpe%
\BCBL {}\ \BBA {} Brunet%
}{%
Bauer%
\ \protect \BOthers {.}}{%
{\protect \APACyear {2015}}%
}]{%
bauer2015quiet}
\APACinsertmetastar {%
bauer2015quiet}%
\begin{APACrefauthors}%
Bauer, P.%
, Thorpe, A.%
\BCBL {}\ \BBA {} Brunet, G.%
\end{APACrefauthors}%
\unskip\
\newblock
\APACrefYearMonthDay{2015}{}{}.
\newblock
{\BBOQ}\APACrefatitle {The quiet revolution of numerical weather prediction} {The quiet revolution of numerical weather prediction}.{\BBCQ}
\newblock
\APACjournalVolNumPages{Nature}{525}{7567}{47--55}.
\PrintBackRefs{\CurrentBib}

\bibitem [\protect \citeauthoryear {%
Bengio%
\ \protect \BOthers {.}}{%
Bengio%
\ \protect \BOthers {.}}{%
{\protect \APACyear {2009}}%
}]{%
bengio2009learning}
\APACinsertmetastar {%
bengio2009learning}%
\begin{APACrefauthors}%
Bengio, Y.%
\BCBT {}\ \BOthersPeriod {.}
\end{APACrefauthors}%
\unskip\
\newblock
\APACrefYearMonthDay{2009}{}{}.
\newblock
{\BBOQ}\APACrefatitle {Learning deep architectures for AI} {Learning deep architectures for ai}.{\BBCQ}
\newblock
\APACjournalVolNumPages{Foundations and trends{\textregistered} in Machine Learning}{2}{1}{1--127}.
\PrintBackRefs{\CurrentBib}

\bibitem [\protect \citeauthoryear {%
Berner%
\ \protect \BOthers {.}}{%
Berner%
\ \protect \BOthers {.}}{%
{\protect \APACyear {2017}}%
}]{%
berner2017stochastic}
\APACinsertmetastar {%
berner2017stochastic}%
\begin{APACrefauthors}%
Berner, J.%
, Achatz, U.%
, Batte, L.%
, Bengtsson, L.%
, C{\'a}mara, A\BPBI d\BPBI l.%
, Christensen, H\BPBI M.%
\BDBL {}others%
\end{APACrefauthors}%
\unskip\
\newblock
\APACrefYearMonthDay{2017}{}{}.
\newblock
{\BBOQ}\APACrefatitle {Stochastic parameterization: Toward a new view of weather and climate models} {Stochastic parameterization: Toward a new view of weather and climate models}.{\BBCQ}
\newblock
\APACjournalVolNumPages{Bulletin of the American Meteorological Society}{98}{3}{565--588}.
\PrintBackRefs{\CurrentBib}

\bibitem [\protect \citeauthoryear {%
Bi%
\ \protect \BOthers {.}}{%
Bi%
\ \protect \BOthers {.}}{%
{\protect \APACyear {2023}}%
}]{%
bi2023accurate}
\APACinsertmetastar {%
bi2023accurate}%
\begin{APACrefauthors}%
Bi, K.%
, Xie, L.%
, Zhang, H.%
, Chen, X.%
, Gu, X.%
\BCBL {}\ \BBA {} Tian, Q.%
\end{APACrefauthors}%
\unskip\
\newblock
\APACrefYearMonthDay{2023}{}{}.
\newblock
{\BBOQ}\APACrefatitle {Accurate medium-range global weather forecasting with 3D neural networks} {Accurate medium-range global weather forecasting with 3d neural networks}.{\BBCQ}
\newblock
\APACjournalVolNumPages{Nature}{619}{7970}{533--538}.
\PrintBackRefs{\CurrentBib}

\bibitem [\protect \citeauthoryear {%
Br{\'a}s%
, Simoes%
, Amorim%
\BCBL {}\ \BBA {} Fortes%
}{%
Br{\'a}s%
\ \protect \BOthers {.}}{%
{\protect \APACyear {2023}}%
}]{%
bras2023much}
\APACinsertmetastar {%
bras2023much}%
\begin{APACrefauthors}%
Br{\'a}s, T\BPBI A.%
, Simoes, S.%
, Amorim, F.%
\BCBL {}\ \BBA {} Fortes, P.%
\end{APACrefauthors}%
\unskip\
\newblock
\APACrefYearMonthDay{2023}{}{}.
\newblock
{\BBOQ}\APACrefatitle {How much extreme weather events have affected European power generation in the past three decades?} {How much extreme weather events have affected european power generation in the past three decades?}{\BBCQ}
\newblock
\APACjournalVolNumPages{Renewable and Sustainable Energy Reviews}{183}{}{113494}.
\PrintBackRefs{\CurrentBib}

\bibitem [\protect \citeauthoryear {%
Brenowitz%
\ \BBA {} Bretherton%
}{%
Brenowitz%
\ \BBA {} Bretherton%
}{%
{\protect \APACyear {2018}}%
}]{%
brenowitz2018prognostic}
\APACinsertmetastar {%
brenowitz2018prognostic}%
\begin{APACrefauthors}%
Brenowitz, N\BPBI D.%
\BCBT {}\ \BBA {} Bretherton, C\BPBI S.%
\end{APACrefauthors}%
\unskip\
\newblock
\APACrefYearMonthDay{2018}{}{}.
\newblock
{\BBOQ}\APACrefatitle {Prognostic validation of a neural network unified physics parameterization} {Prognostic validation of a neural network unified physics parameterization}.{\BBCQ}
\newblock
\APACjournalVolNumPages{Geophysical Research Letters}{45}{12}{6289--6298}.
\PrintBackRefs{\CurrentBib}

\bibitem [\protect \citeauthoryear {%
Cao%
, Zhang%
, Lv%
, Yu%
\BCBL {}\ \BBA {} Ai%
}{%
Cao%
\ \protect \BOthers {.}}{%
{\protect \APACyear {2025}}%
}]{%
cao2025ai}
\APACinsertmetastar {%
cao2025ai}%
\begin{APACrefauthors}%
Cao, Y.%
, Zhang, S.%
, Lv, G.%
, Yu, M.%
\BCBL {}\ \BBA {} Ai, B.%
\end{APACrefauthors}%
\unskip\
\newblock
\APACrefYearMonthDay{2025}{}{}.
\newblock
{\BBOQ}\APACrefatitle {AI-based Correction of Wave Forecasts Using the Transformer-enhanced UNet Model} {Ai-based correction of wave forecasts using the transformer-enhanced unet model}.{\BBCQ}
\newblock
\APACjournalVolNumPages{Advances in Atmospheric Sciences}{42}{1}{221--231}.
\PrintBackRefs{\CurrentBib}

\bibitem [\protect \citeauthoryear {%
Chandramouli%
, Wang%
, Johnson%
\BCBL {}\ \BBA {} Otkin%
}{%
Chandramouli%
\ \protect \BOthers {.}}{%
{\protect \APACyear {2022}}%
}]{%
chandramouli2022online}
\APACinsertmetastar {%
chandramouli2022online}%
\begin{APACrefauthors}%
Chandramouli, K.%
, Wang, X.%
, Johnson, A.%
\BCBL {}\ \BBA {} Otkin, J.%
\end{APACrefauthors}%
\unskip\
\newblock
\APACrefYearMonthDay{2022}{}{}.
\newblock
{\BBOQ}\APACrefatitle {Online nonlinear bias correction in ensemble Kalman filter to assimilate GOES-R all-sky radiances for the analysis and prediction of rapidly developing supercells} {Online nonlinear bias correction in ensemble kalman filter to assimilate goes-r all-sky radiances for the analysis and prediction of rapidly developing supercells}.{\BBCQ}
\newblock
\APACjournalVolNumPages{Journal of Advances in Modeling Earth Systems}{14}{3}{e2021MS002711}.
\PrintBackRefs{\CurrentBib}

\bibitem [\protect \citeauthoryear {%
Chassignet%
\ \protect \BOthers {.}}{%
Chassignet%
\ \protect \BOthers {.}}{%
{\protect \APACyear {2007}}%
}]{%
chassignet2007hycom}
\APACinsertmetastar {%
chassignet2007hycom}%
\begin{APACrefauthors}%
Chassignet, E\BPBI P.%
, Hurlburt, H\BPBI E.%
, Smedstad, O\BPBI M.%
, Halliwell, G\BPBI R.%
, Hogan, P\BPBI J.%
, Wallcraft, A\BPBI J.%
\BDBL {}Bleck, R.%
\end{APACrefauthors}%
\unskip\
\newblock
\APACrefYearMonthDay{2007}{}{}.
\newblock
{\BBOQ}\APACrefatitle {The HYCOM (hybrid coordinate ocean model) data assimilative system} {The hycom (hybrid coordinate ocean model) data assimilative system}.{\BBCQ}
\newblock
\APACjournalVolNumPages{Journal of Marine Systems}{65}{1-4}{60--83}.
\PrintBackRefs{\CurrentBib}

\bibitem [\protect \citeauthoryear {%
Dey%
\ \BBA {} Salem%
}{%
Dey%
\ \BBA {} Salem%
}{%
{\protect \APACyear {2017}}%
}]{%
dey2017gate}
\APACinsertmetastar {%
dey2017gate}%
\begin{APACrefauthors}%
Dey, R.%
\BCBT {}\ \BBA {} Salem, F\BPBI M.%
\end{APACrefauthors}%
\unskip\
\newblock
\APACrefYearMonthDay{2017}{}{}.
\newblock
{\BBOQ}\APACrefatitle {Gate-variants of gated recurrent unit (GRU) neural networks} {Gate-variants of gated recurrent unit (gru) neural networks}.{\BBCQ}
\newblock
\BIn{} \APACrefbtitle {2017 IEEE 60th international midwest symposium on circuits and systems (MWSCAS)} {2017 ieee 60th international midwest symposium on circuits and systems (mwscas)}\ (\BPGS\ 1597--1600).
\PrintBackRefs{\CurrentBib}

\bibitem [\protect \citeauthoryear {%
Fang%
\ \protect \BOthers {.}}{%
Fang%
\ \protect \BOthers {.}}{%
{\protect \APACyear {2023}}%
}]{%
fang2023short}
\APACinsertmetastar {%
fang2023short}%
\begin{APACrefauthors}%
Fang, Y.%
, Wu, Y.%
, Wu, F.%
, Yan, Y.%
, Liu, Q.%
, Liu, N.%
\BCBL {}\ \BBA {} Xia, J.%
\end{APACrefauthors}%
\unskip\
\newblock
\APACrefYearMonthDay{2023}{}{}.
\newblock
{\BBOQ}\APACrefatitle {Short-term wind speed forecasting bias correction in the Hangzhou area of China based on a machine learning model} {Short-term wind speed forecasting bias correction in the hangzhou area of china based on a machine learning model}.{\BBCQ}
\newblock
\APACjournalVolNumPages{Atmospheric and Oceanic Science Letters}{16}{4}{100339}.
\PrintBackRefs{\CurrentBib}

\bibitem [\protect \citeauthoryear {%
Glorot%
\ \BBA {} Bengio%
}{%
Glorot%
\ \BBA {} Bengio%
}{%
{\protect \APACyear {2010}}%
}]{%
glorot2010understanding}
\APACinsertmetastar {%
glorot2010understanding}%
\begin{APACrefauthors}%
Glorot, X.%
\BCBT {}\ \BBA {} Bengio, Y.%
\end{APACrefauthors}%
\unskip\
\newblock
\APACrefYearMonthDay{2010}{}{}.
\newblock
{\BBOQ}\APACrefatitle {Understanding the difficulty of training deep feedforward neural networks} {Understanding the difficulty of training deep feedforward neural networks}.{\BBCQ}
\newblock
\BIn{} \APACrefbtitle {Proceedings of the thirteenth international conference on artificial intelligence and statistics} {Proceedings of the thirteenth international conference on artificial intelligence and statistics}\ (\BPGS\ 249--256).
\PrintBackRefs{\CurrentBib}

\bibitem [\protect \citeauthoryear {%
Goodfellow%
\ \protect \BOthers {.}}{%
Goodfellow%
\ \protect \BOthers {.}}{%
{\protect \APACyear {2020}}%
}]{%
goodfellow2020generative}
\APACinsertmetastar {%
goodfellow2020generative}%
\begin{APACrefauthors}%
Goodfellow, I.%
, Pouget-Abadie, J.%
, Mirza, M.%
, Xu, B.%
, Warde-Farley, D.%
, Ozair, S.%
\BDBL {}Bengio, Y.%
\end{APACrefauthors}%
\unskip\
\newblock
\APACrefYearMonthDay{2020}{}{}.
\newblock
{\BBOQ}\APACrefatitle {Generative adversarial networks} {Generative adversarial networks}.{\BBCQ}
\newblock
\APACjournalVolNumPages{Communications of the ACM}{63}{11}{139--144}.
\PrintBackRefs{\CurrentBib}

\bibitem [\protect \citeauthoryear {%
Han%
\ \protect \BOthers {.}}{%
Han%
\ \protect \BOthers {.}}{%
{\protect \APACyear {2021}}%
}]{%
han2021deep}
\APACinsertmetastar {%
han2021deep}%
\begin{APACrefauthors}%
Han, L.%
, Chen, M.%
, Chen, K.%
, Chen, H.%
, Zhang, Y.%
, Lu, B.%
\BDBL {}Qin, R.%
\end{APACrefauthors}%
\unskip\
\newblock
\APACrefYearMonthDay{2021}{}{}.
\newblock
{\BBOQ}\APACrefatitle {A deep learning method for bias correction of ECMWF 24--240 h forecasts} {A deep learning method for bias correction of ecmwf 24--240 h forecasts}.{\BBCQ}
\newblock
\APACjournalVolNumPages{Advances in Atmospheric Sciences}{38}{9}{1444--1459}.
\PrintBackRefs{\CurrentBib}

\bibitem [\protect \citeauthoryear {%
He%
, Zhang%
, Ren%
\BCBL {}\ \BBA {} Sun%
}{%
He%
\ \protect \BOthers {.}}{%
{\protect \APACyear {2016}}%
}]{%
he2016deep}
\APACinsertmetastar {%
he2016deep}%
\begin{APACrefauthors}%
He, K.%
, Zhang, X.%
, Ren, S.%
\BCBL {}\ \BBA {} Sun, J.%
\end{APACrefauthors}%
\unskip\
\newblock
\APACrefYearMonthDay{2016}{}{}.
\newblock
{\BBOQ}\APACrefatitle {Deep residual learning for image recognition} {Deep residual learning for image recognition}.{\BBCQ}
\newblock
\BIn{} \APACrefbtitle {Proceedings of the IEEE conference on computer vision and pattern recognition} {Proceedings of the ieee conference on computer vision and pattern recognition}\ (\BPGS\ 770--778).
\PrintBackRefs{\CurrentBib}

\bibitem [\protect \citeauthoryear {%
Hersbach%
\ \protect \BOthers {.}}{%
Hersbach%
\ \protect \BOthers {.}}{%
{\protect \APACyear {2020}}%
}]{%
hersbach2020era5}
\APACinsertmetastar {%
hersbach2020era5}%
\begin{APACrefauthors}%
Hersbach, H.%
, Bell, B.%
, Berrisford, P.%
, Hirahara, S.%
, Hor{\'a}nyi, A.%
, Mu{\~n}oz-Sabater, J.%
\BDBL {}others%
\end{APACrefauthors}%
\unskip\
\newblock
\APACrefYearMonthDay{2020}{}{}.
\newblock
{\BBOQ}\APACrefatitle {The ERA5 global reanalysis} {The era5 global reanalysis}.{\BBCQ}
\newblock
\APACjournalVolNumPages{Quarterly journal of the royal meteorological society}{146}{730}{1999--2049}.
\PrintBackRefs{\CurrentBib}

\bibitem [\protect \citeauthoryear {%
Ioffe%
\ \BBA {} Szegedy%
}{%
Ioffe%
\ \BBA {} Szegedy%
}{%
{\protect \APACyear {2015}}%
}]{%
ioffe2015batch}
\APACinsertmetastar {%
ioffe2015batch}%
\begin{APACrefauthors}%
Ioffe, S.%
\BCBT {}\ \BBA {} Szegedy, C.%
\end{APACrefauthors}%
\unskip\
\newblock
\APACrefYearMonthDay{2015}{}{}.
\newblock
{\BBOQ}\APACrefatitle {Batch normalization: Accelerating deep network training by reducing internal covariate shift} {Batch normalization: Accelerating deep network training by reducing internal covariate shift}.{\BBCQ}
\newblock
\BIn{} \APACrefbtitle {International conference on machine learning} {International conference on machine learning}\ (\BPGS\ 448--456).
\PrintBackRefs{\CurrentBib}

\bibitem [\protect \citeauthoryear {%
Johnson%
\ \protect \BOthers {.}}{%
Johnson%
\ \protect \BOthers {.}}{%
{\protect \APACyear {2019}}%
}]{%
johnson2019seas5}
\APACinsertmetastar {%
johnson2019seas5}%
\begin{APACrefauthors}%
Johnson, S\BPBI J.%
, Stockdale, T\BPBI N.%
, Ferranti, L.%
, Balmaseda, M\BPBI A.%
, Molteni, F.%
, Magnusson, L.%
\BDBL {}others%
\end{APACrefauthors}%
\unskip\
\newblock
\APACrefYearMonthDay{2019}{}{}.
\newblock
{\BBOQ}\APACrefatitle {SEAS5: the new ECMWF seasonal forecast system} {Seas5: the new ecmwf seasonal forecast system}.{\BBCQ}
\newblock
\APACjournalVolNumPages{Geoscientific Model Development}{12}{3}{1087--1117}.
\PrintBackRefs{\CurrentBib}

\bibitem [\protect \citeauthoryear {%
LeCun%
, Bottou%
, Bengio%
\BCBL {}\ \BBA {} Haffner%
}{%
LeCun%
\ \protect \BOthers {.}}{%
{\protect \APACyear {1998}}%
}]{%
lecun1998gradient}
\APACinsertmetastar {%
lecun1998gradient}%
\begin{APACrefauthors}%
LeCun, Y.%
, Bottou, L.%
, Bengio, Y.%
\BCBL {}\ \BBA {} Haffner, P.%
\end{APACrefauthors}%
\unskip\
\newblock
\APACrefYearMonthDay{1998}{}{}.
\newblock
{\BBOQ}\APACrefatitle {Gradient-based learning applied to document recognition} {Gradient-based learning applied to document recognition}.{\BBCQ}
\newblock
\APACjournalVolNumPages{Proceedings of the IEEE}{86}{11}{2278--2324}.
\PrintBackRefs{\CurrentBib}

\bibitem [\protect \citeauthoryear {%
Lin%
, Li%
, Zheng%
, Cheng%
\BCBL {}\ \BBA {} Yuan%
}{%
Lin%
\ \protect \BOthers {.}}{%
{\protect \APACyear {2020}}%
}]{%
lin2020self}
\APACinsertmetastar {%
lin2020self}%
\begin{APACrefauthors}%
Lin, Z.%
, Li, M.%
, Zheng, Z.%
, Cheng, Y.%
\BCBL {}\ \BBA {} Yuan, C.%
\end{APACrefauthors}%
\unskip\
\newblock
\APACrefYearMonthDay{2020}{}{}.
\newblock
{\BBOQ}\APACrefatitle {Self-attention convlstm for spatiotemporal prediction} {Self-attention convlstm for spatiotemporal prediction}.{\BBCQ}
\newblock
\BIn{} \APACrefbtitle {Proceedings of the AAAI conference on artificial intelligence} {Proceedings of the aaai conference on artificial intelligence}\ (\BVOL~34, \BPGS\ 11531--11538).
\PrintBackRefs{\CurrentBib}

\bibitem [\protect \citeauthoryear {%
McGibbon%
\ \protect \BOthers {.}}{%
McGibbon%
\ \protect \BOthers {.}}{%
{\protect \APACyear {2024}}%
}]{%
mcgibbon2024global}
\APACinsertmetastar {%
mcgibbon2024global}%
\begin{APACrefauthors}%
McGibbon, J.%
, Clark, S.%
, Henn, B.%
, Kwa, A.%
, Watt-Meyer, O.%
, Perkins, W.%
\BCBL {}\ \BBA {} Bretherton, C.%
\end{APACrefauthors}%
\unskip\
\newblock
\APACrefYearMonthDay{2024}{}{}.
\newblock
{\BBOQ}\APACrefatitle {Global precipitation correction across a range of climates using CycleGAN} {Global precipitation correction across a range of climates using cyclegan}.{\BBCQ}
\newblock
\APACjournalVolNumPages{Geophysical Research Letters}{51}{4}{e2023GL105131}.
\PrintBackRefs{\CurrentBib}

\bibitem [\protect \citeauthoryear {%
Miao%
\ \protect \BOthers {.}}{%
Miao%
\ \protect \BOthers {.}}{%
{\protect \APACyear {2024}}%
}]{%
miao2024unveiling}
\APACinsertmetastar {%
miao2024unveiling}%
\begin{APACrefauthors}%
Miao, L.%
, Ju, L.%
, Sun, S.%
, Agathokleous, E.%
, Wang, Q.%
, Zhu, Z.%
\BDBL {}Liu, Q.%
\end{APACrefauthors}%
\unskip\
\newblock
\APACrefYearMonthDay{2024}{}{}.
\newblock
{\BBOQ}\APACrefatitle {Unveiling the dynamics of sequential extreme precipitation-heatwave compounds in China} {Unveiling the dynamics of sequential extreme precipitation-heatwave compounds in china}.{\BBCQ}
\newblock
\APACjournalVolNumPages{npj Climate and Atmospheric Science}{7}{1}{67}.
\PrintBackRefs{\CurrentBib}

\bibitem [\protect \citeauthoryear {%
Mishra%
\ \protect \BOthers {.}}{%
Mishra%
\ \protect \BOthers {.}}{%
{\protect \APACyear {2021}}%
}]{%
mishra2021impact}
\APACinsertmetastar {%
mishra2021impact}%
\begin{APACrefauthors}%
Mishra, A\BPBI K.%
, Kumar, P.%
, Dubey, A\BPBI K.%
, Javed, A.%
, Saharwardi, M\BPBI S.%
, Sein, D\BPBI V.%
\BDBL {}Jacob, D.%
\end{APACrefauthors}%
\unskip\
\newblock
\APACrefYearMonthDay{2021}{}{}.
\newblock
{\BBOQ}\APACrefatitle {Impact of horizontal resolution on monsoon precipitation for CORDEX-South Asia: a regional earth system model assessment} {Impact of horizontal resolution on monsoon precipitation for cordex-south asia: a regional earth system model assessment}.{\BBCQ}
\newblock
\APACjournalVolNumPages{Atmospheric Research}{259}{}{105681}.
\PrintBackRefs{\CurrentBib}

\bibitem [\protect \citeauthoryear {%
Molteni%
, Buizza%
, Palmer%
\BCBL {}\ \BBA {} Petroliagis%
}{%
Molteni%
\ \protect \BOthers {.}}{%
{\protect \APACyear {1996}}%
}]{%
molteni1996ecmwf}
\APACinsertmetastar {%
molteni1996ecmwf}%
\begin{APACrefauthors}%
Molteni, F.%
, Buizza, R.%
, Palmer, T\BPBI N.%
\BCBL {}\ \BBA {} Petroliagis, T.%
\end{APACrefauthors}%
\unskip\
\newblock
\APACrefYearMonthDay{1996}{}{}.
\newblock
{\BBOQ}\APACrefatitle {The ECMWF ensemble prediction system: Methodology and validation} {The ecmwf ensemble prediction system: Methodology and validation}.{\BBCQ}
\newblock
\APACjournalVolNumPages{Quarterly journal of the royal meteorological society}{122}{529}{73--119}.
\PrintBackRefs{\CurrentBib}

\bibitem [\protect \citeauthoryear {%
Pathak%
\ \protect \BOthers {.}}{%
Pathak%
\ \protect \BOthers {.}}{%
{\protect \APACyear {2022}}%
}]{%
pathak2022fourcastnet}
\APACinsertmetastar {%
pathak2022fourcastnet}%
\begin{APACrefauthors}%
Pathak, J.%
, Subramanian, S.%
, Harrington, P.%
, Raja, S.%
, Chattopadhyay, A.%
, Mardani, M.%
\BDBL {}others%
\end{APACrefauthors}%
\unskip\
\newblock
\APACrefYearMonthDay{2022}{}{}.
\newblock
{\BBOQ}\APACrefatitle {Fourcastnet: A global data-driven high-resolution weather model using adaptive fourier neural operators} {Fourcastnet: A global data-driven high-resolution weather model using adaptive fourier neural operators}.{\BBCQ}
\newblock
\APACjournalVolNumPages{arXiv preprint arXiv:2202.11214}{}{}{}.
\PrintBackRefs{\CurrentBib}

\bibitem [\protect \citeauthoryear {%
Peng%
\ \BBA {} Xie%
}{%
Peng%
\ \BBA {} Xie%
}{%
{\protect \APACyear {2006}}%
}]{%
peng2006effect}
\APACinsertmetastar {%
peng2006effect}%
\begin{APACrefauthors}%
Peng, S\BHBI Q.%
\BCBT {}\ \BBA {} Xie, L.%
\end{APACrefauthors}%
\unskip\
\newblock
\APACrefYearMonthDay{2006}{}{}.
\newblock
{\BBOQ}\APACrefatitle {Effect of determining initial conditions by four-dimensional variational data assimilation on storm surge forecasting} {Effect of determining initial conditions by four-dimensional variational data assimilation on storm surge forecasting}.{\BBCQ}
\newblock
\APACjournalVolNumPages{Ocean Modelling}{14}{1-2}{1--18}.
\PrintBackRefs{\CurrentBib}

\bibitem [\protect \citeauthoryear {%
Ronneberger%
, Fischer%
\BCBL {}\ \BBA {} Brox%
}{%
Ronneberger%
\ \protect \BOthers {.}}{%
{\protect \APACyear {2015}}%
}]{%
ronneberger2015u}
\APACinsertmetastar {%
ronneberger2015u}%
\begin{APACrefauthors}%
Ronneberger, O.%
, Fischer, P.%
\BCBL {}\ \BBA {} Brox, T.%
\end{APACrefauthors}%
\unskip\
\newblock
\APACrefYearMonthDay{2015}{}{}.
\newblock
{\BBOQ}\APACrefatitle {U-net: Convolutional networks for biomedical image segmentation} {U-net: Convolutional networks for biomedical image segmentation}.{\BBCQ}
\newblock
\BIn{} \APACrefbtitle {Medical image computing and computer-assisted intervention--MICCAI 2015: 18th international conference, Munich, Germany, October 5-9, 2015, proceedings, part III 18} {Medical image computing and computer-assisted intervention--miccai 2015: 18th international conference, munich, germany, october 5-9, 2015, proceedings, part iii 18}\ (\BPGS\ 234--241).
\PrintBackRefs{\CurrentBib}

\bibitem [\protect \citeauthoryear {%
Shi%
\ \protect \BOthers {.}}{%
Shi%
\ \protect \BOthers {.}}{%
{\protect \APACyear {2015}}%
}]{%
shi2015convolutional}
\APACinsertmetastar {%
shi2015convolutional}%
\begin{APACrefauthors}%
Shi, X.%
, Chen, Z.%
, Wang, H.%
, Yeung, D\BHBI Y.%
, Wong, W\BHBI K.%
\BCBL {}\ \BBA {} Woo, W\BHBI c.%
\end{APACrefauthors}%
\unskip\
\newblock
\APACrefYearMonthDay{2015}{}{}.
\newblock
{\BBOQ}\APACrefatitle {Convolutional LSTM network: A machine learning approach for precipitation nowcasting} {Convolutional lstm network: A machine learning approach for precipitation nowcasting}.{\BBCQ}
\newblock
\APACjournalVolNumPages{Advances in neural information processing systems}{28}{}{}.
\PrintBackRefs{\CurrentBib}

\bibitem [\protect \citeauthoryear {%
Tishby%
\ \BBA {} Zaslavsky%
}{%
Tishby%
\ \BBA {} Zaslavsky%
}{%
{\protect \APACyear {2015}}%
}]{%
tishby2015deep}
\APACinsertmetastar {%
tishby2015deep}%
\begin{APACrefauthors}%
Tishby, N.%
\BCBT {}\ \BBA {} Zaslavsky, N.%
\end{APACrefauthors}%
\unskip\
\newblock
\APACrefYearMonthDay{2015}{}{}.
\newblock
{\BBOQ}\APACrefatitle {Deep learning and the information bottleneck principle} {Deep learning and the information bottleneck principle}.{\BBCQ}
\newblock
\BIn{} \APACrefbtitle {2015 ieee information theory workshop (itw)} {2015 ieee information theory workshop (itw)}\ (\BPGS\ 1--5).
\PrintBackRefs{\CurrentBib}

\bibitem [\protect \citeauthoryear {%
Turco%
, Llasat%
, Herrera%
\BCBL {}\ \BBA {} Guti{\'e}rrez%
}{%
Turco%
\ \protect \BOthers {.}}{%
{\protect \APACyear {2017}}%
}]{%
turco2017bias}
\APACinsertmetastar {%
turco2017bias}%
\begin{APACrefauthors}%
Turco, M.%
, Llasat, M\BPBI C.%
, Herrera, S.%
\BCBL {}\ \BBA {} Guti{\'e}rrez, J\BPBI M.%
\end{APACrefauthors}%
\unskip\
\newblock
\APACrefYearMonthDay{2017}{}{}.
\newblock
{\BBOQ}\APACrefatitle {Bias correction and downscaling of future RCM precipitation projections using a MOS-Analog technique} {Bias correction and downscaling of future rcm precipitation projections using a mos-analog technique}.{\BBCQ}
\newblock
\APACjournalVolNumPages{Journal of Geophysical Research: Atmospheres}{122}{5}{2631--2648}.
\PrintBackRefs{\CurrentBib}

\bibitem [\protect \citeauthoryear {%
Wang%
, Tian%
\BCBL {}\ \BBA {} Carroll%
}{%
Wang%
\ \protect \BOthers {.}}{%
{\protect \APACyear {2022}}%
}]{%
wang2022customized}
\APACinsertmetastar {%
wang2022customized}%
\begin{APACrefauthors}%
Wang, F.%
, Tian, D.%
\BCBL {}\ \BBA {} Carroll, M.%
\end{APACrefauthors}%
\unskip\
\newblock
\APACrefYearMonthDay{2022}{}{}.
\newblock
{\BBOQ}\APACrefatitle {Customized deep learning for precipitation bias correction and downscaling} {Customized deep learning for precipitation bias correction and downscaling}.{\BBCQ}
\newblock
\APACjournalVolNumPages{Geoscientific Model Development Discussions}{2022}{}{1--38}.
\PrintBackRefs{\CurrentBib}

\bibitem [\protect \citeauthoryear {%
Xiong%
\ \protect \BOthers {.}}{%
Xiong%
\ \protect \BOthers {.}}{%
{\protect \APACyear {2023}}%
}]{%
xiong2023ai}
\APACinsertmetastar {%
xiong2023ai}%
\begin{APACrefauthors}%
Xiong, W.%
, Xiang, Y.%
, Wu, H.%
, Zhou, S.%
, Sun, Y.%
, Ma, M.%
\BCBL {}\ \BBA {} Huang, X.%
\end{APACrefauthors}%
\unskip\
\newblock
\APACrefYearMonthDay{2023}{}{}.
\newblock
{\BBOQ}\APACrefatitle {Ai-goms: Large ai-driven global ocean modeling system} {Ai-goms: Large ai-driven global ocean modeling system}.{\BBCQ}
\newblock
\APACjournalVolNumPages{arXiv preprint arXiv:2308.03152}{}{}{}.
\PrintBackRefs{\CurrentBib}

\bibitem [\protect \citeauthoryear {%
Yumnam%
, Guntu%
, Rathinasamy%
\BCBL {}\ \BBA {} Agarwal%
}{%
Yumnam%
\ \protect \BOthers {.}}{%
{\protect \APACyear {2022}}%
}]{%
yumnam2022quantile}
\APACinsertmetastar {%
yumnam2022quantile}%
\begin{APACrefauthors}%
Yumnam, K.%
, Guntu, R\BPBI K.%
, Rathinasamy, M.%
\BCBL {}\ \BBA {} Agarwal, A.%
\end{APACrefauthors}%
\unskip\
\newblock
\APACrefYearMonthDay{2022}{}{}.
\newblock
{\BBOQ}\APACrefatitle {Quantile-based Bayesian Model Averaging approach towards merging of precipitation products} {Quantile-based bayesian model averaging approach towards merging of precipitation products}.{\BBCQ}
\newblock
\APACjournalVolNumPages{Journal of Hydrology}{604}{}{127206}.
\PrintBackRefs{\CurrentBib}

\bibitem [\protect \citeauthoryear {%
Zhang%
\ \protect \BOthers {.}}{%
Zhang%
\ \protect \BOthers {.}}{%
{\protect \APACyear {2023}}%
}]{%
zhang2023deep}
\APACinsertmetastar {%
zhang2023deep}%
\begin{APACrefauthors}%
Zhang, W.%
, Jiang, Y.%
, Dong, J.%
, Song, X.%
, Pang, R.%
, Guoan, B.%
\BCBL {}\ \BBA {} Yu, H.%
\end{APACrefauthors}%
\unskip\
\newblock
\APACrefYearMonthDay{2023}{}{}.
\newblock
{\BBOQ}\APACrefatitle {A deep learning method for real-time bias correction of wind field forecasts in the Western North Pacific} {A deep learning method for real-time bias correction of wind field forecasts in the western north pacific}.{\BBCQ}
\newblock
\APACjournalVolNumPages{Atmospheric Research}{284}{}{106586}.
\PrintBackRefs{\CurrentBib}

\end{thebibliography}

\appendix
\section{Equations}

\begin{equation}
\text{RMSE}(c, t) = \sqrt{\frac{1}{IJ} \sum_{i=1}^{I} \sum_{j=1}^{J} (X_{\text{pred}}(t) [c, i, j] - X_{\text{true}}(l) [c, i, j])^2}
\label{eq:rmse}
\end{equation}

\begin{equation}
\text{ACC}(c, t) = \frac{\sum_{i,j}\hat{X}_{\text{pred}}(t) [c, i, j] \hat{X}_{\text{true}}(t) [c, i, j]}{\sqrt{\sum_{i,j} (\hat{X}_{\text{pred}}(t) [c, i, j])^2 \sum_{m,n}  (\hat{X}_{\text{true}}(t) [c, i, j])^2}}
\label{eq:acc}
\end{equation}

\begin{equation}
Z_{ij} = \frac{X_{ij} - \mu}{\sigma}
\label{eq:z-score}
\end{equation}

\begin{equation}
Z_{i,j,t} = \frac{X_{i,j,t} - \mu_{i,j,t}}{\sigma_{i,j,t}}
\label{eq:z-scoreijt}
\end{equation}

\section{Figures}

\begin{figure}
\noindent\includegraphics[width=\textwidth]{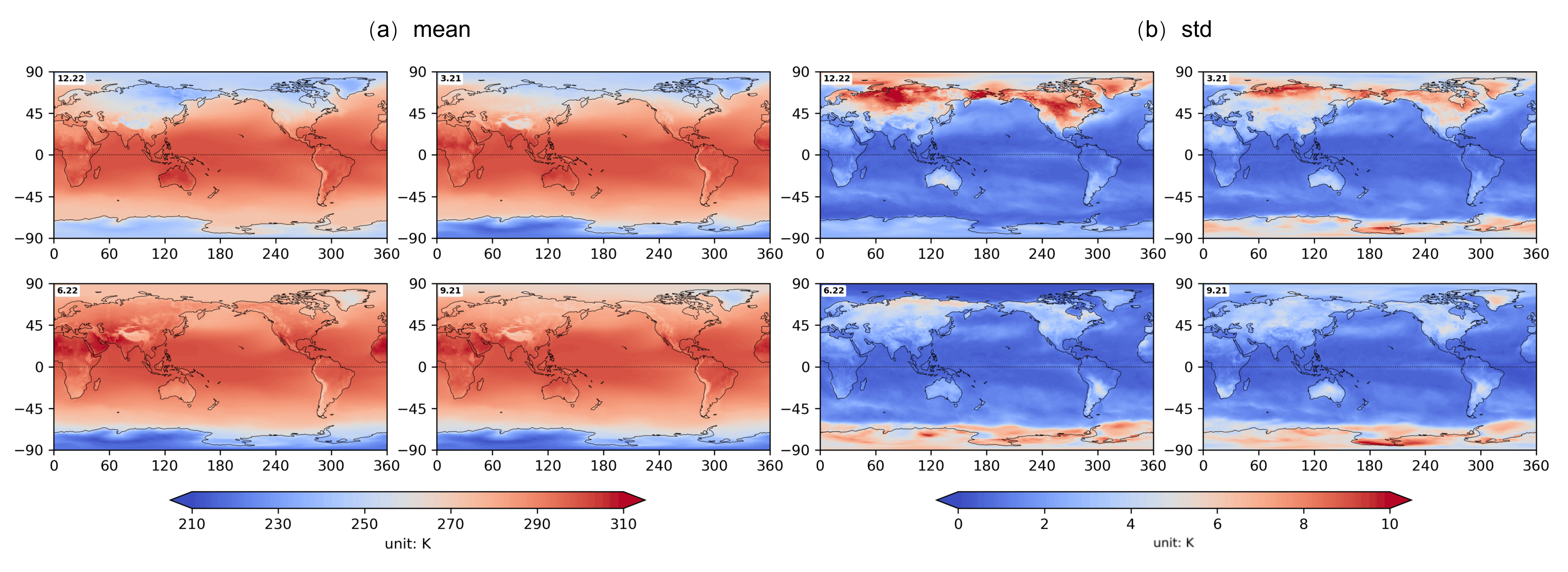}
\caption{The mean and standard deviation (std) representing time points for each season. Taking T2m as an example, the time span is from 1981 to 2021. For each year within this period, we compute the mean and standard deviation for the same day of the year. The representative time points for the four seasons are selected as follows: December 22 for winter, March 21 for spring, June 22 for summer, and September 21 for autumn.}
\label{mean_std}
\end{figure}

\begin{figure}
\noindent\includegraphics[width=\textwidth]{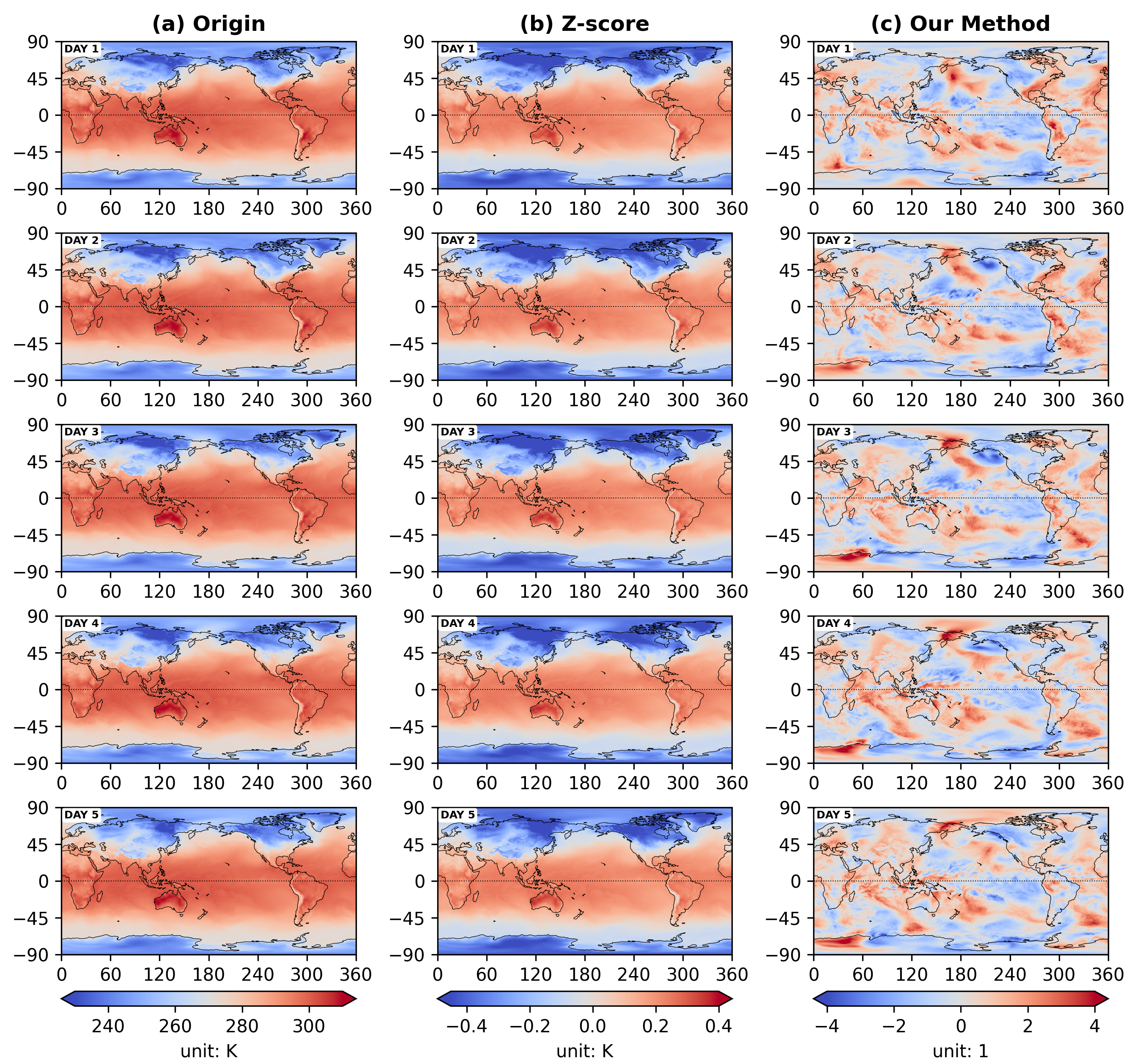}
\caption{True values, Z-score values, our spatiotemporal correlation normalization method values from March 1 to 5, 2021.}
\label{data_orgin_zscore_our}
\end{figure}

\begin{figure}
\noindent\includegraphics[width=\textwidth]{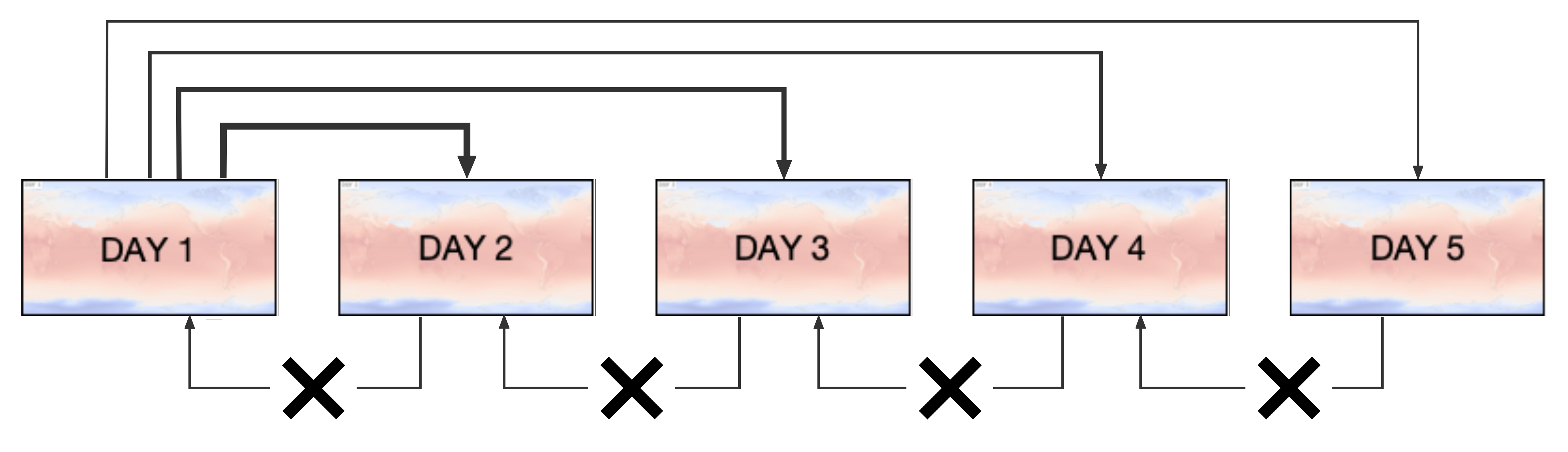}
\caption{The unidirectional time vector diagram. The earlier time steps affect the later ones, with the influence weakening as the time steps progress. However, the later time steps do not affect the earlier ones.}
\label{time_arrow}
\end{figure}

\begin{figure}
\noindent\includegraphics[width=\textwidth]{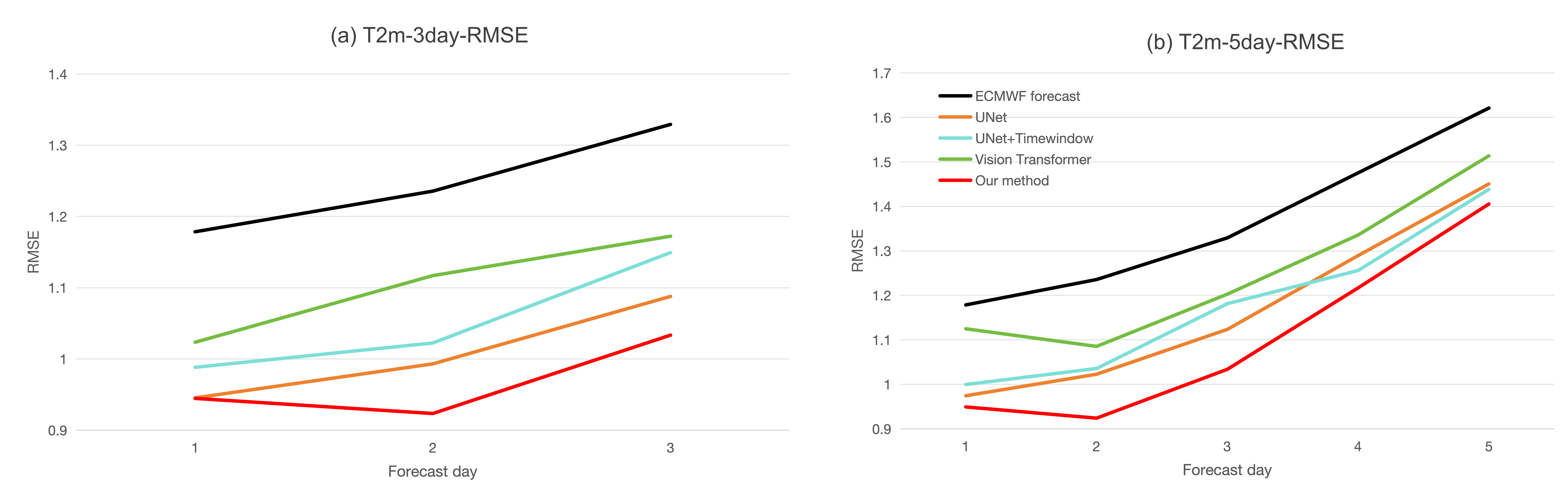}
\caption{The mean correction RMSE for the T2m forecast values from each model of ECMWF starting on the 1st of every month in test set. The left figure illustrates the training correction for a 3-day forecast lead time, while the right figure illustrates the training correction for a 5-day forecast lead time.}
\label{T2m_3day_5day}
\end{figure}

\begin{figure}
\noindent\includegraphics[width=\textwidth]{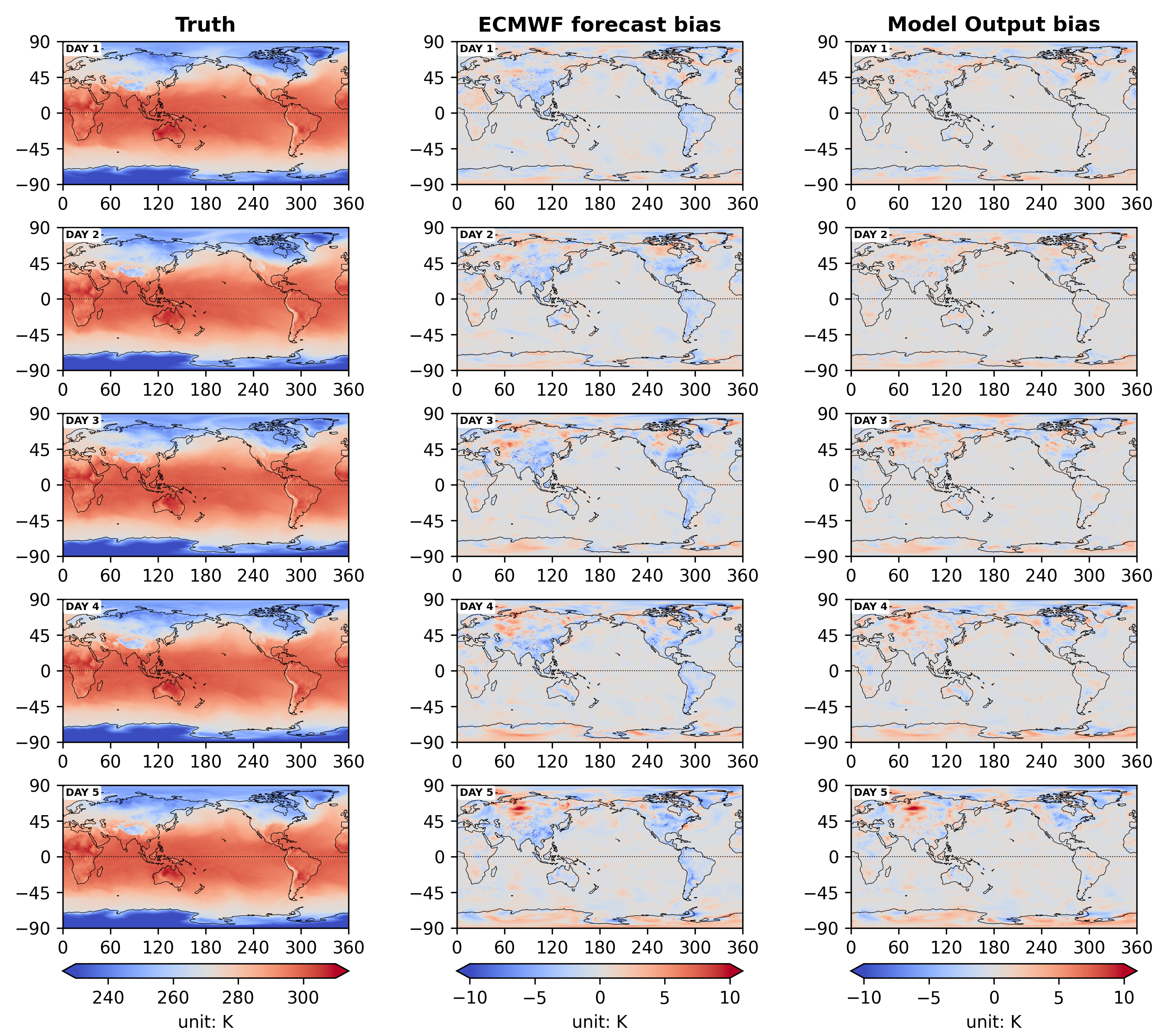}
\caption{The ground truth, the biases of ECMWF forecasts and our model's corrected output relative to ERA5 from March 1 to 5, 2021}
\label{true_ec_model_bias}
\end{figure}

\begin{figure}
\noindent\includegraphics[width=\textwidth]{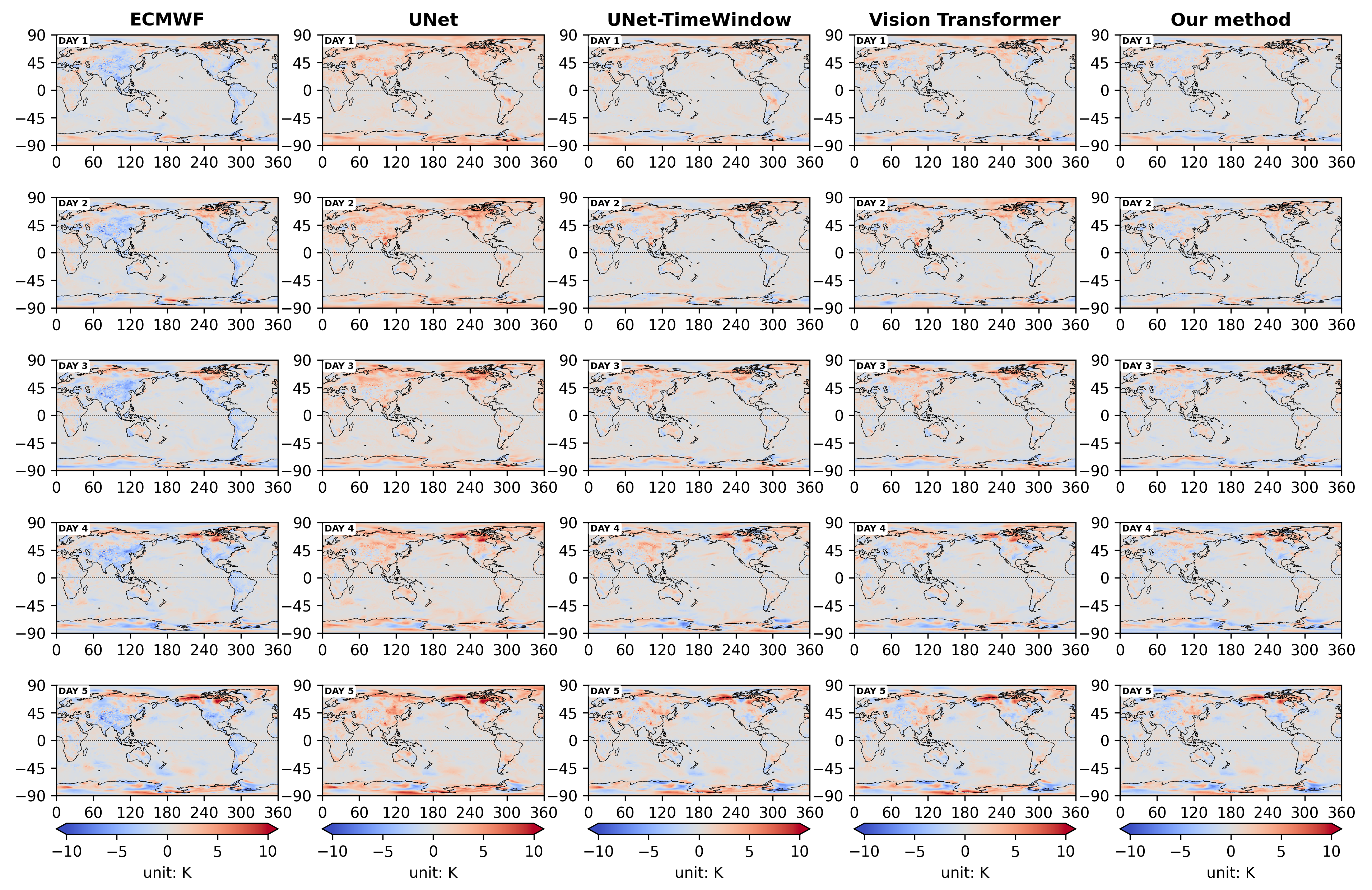}
\caption{The bias of truth between ECMWF forecast values, UNet values, UNet-TimeWindow values, Vision Transformer values, and our method values of T2m from March 1 to 5, 2021.}
\label{ec_unet_vit_our_bias}
\end{figure}

\begin{figure}
\noindent\includegraphics[width=\textwidth]{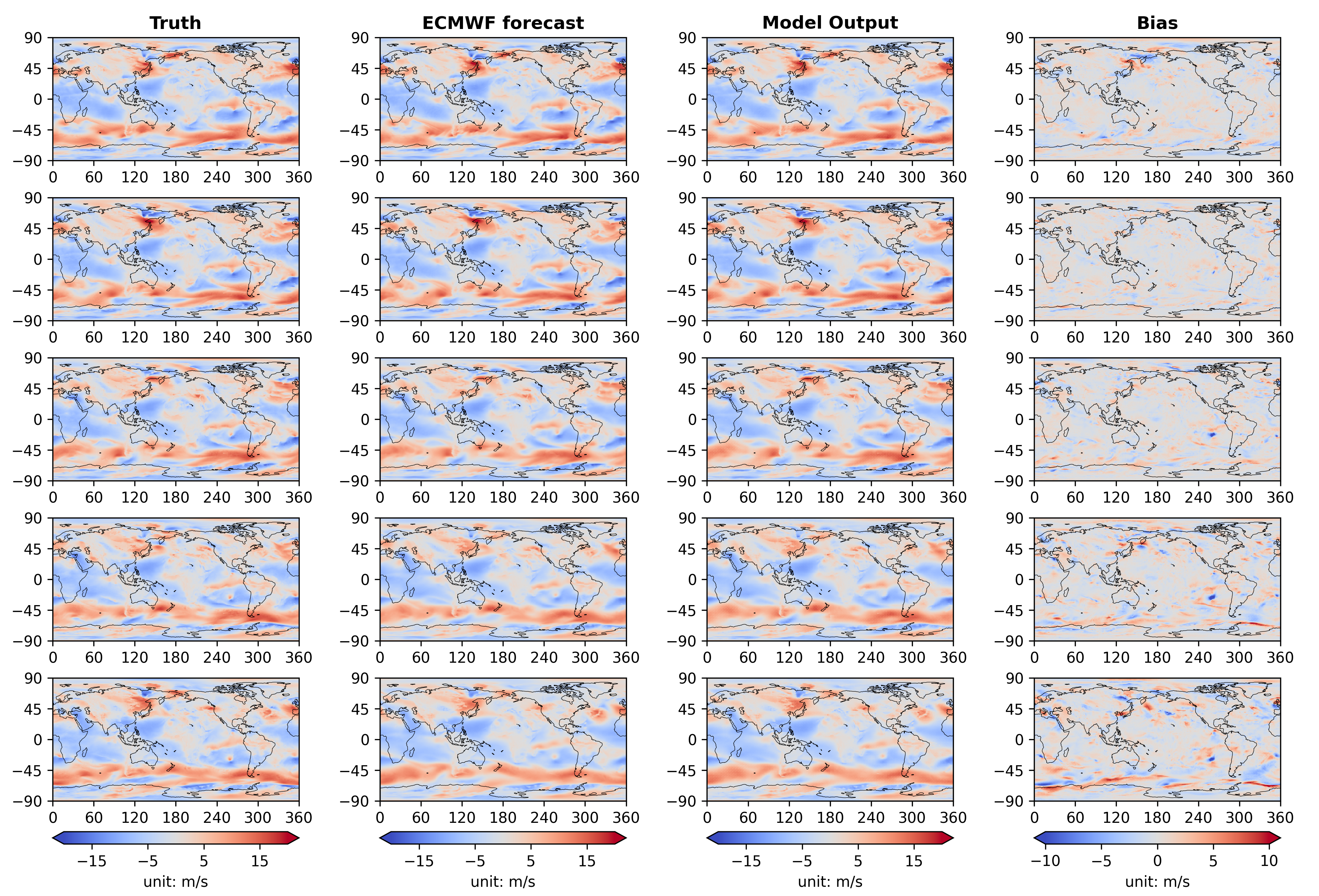}
\caption{The same results as the Figure \ref{true_ec_model_bias}, but with U10 variable.}
\label{u10_true_ec_model_bias}
\end{figure}

\end{document}